\documentclass[letterpaper, 10 pt, conference]{ieeeconf}  

\IEEEoverridecommandlockouts                              
\usepackage{graphicx}
\graphicspath{ {./images/} }

\title{\LARGE \bf
Car Detection using Unmanned Aerial Vehicles: Comparison between Faster R-CNN and YOLOv3
}

\author{Bilel Benjdira$^{1,5}$, Taha Khursheed $^{2}$, Anis Koubaa $^{3}$, Adel Ammar $^{4}$, Kais Ouni$^{5}$
\thanks{*This work is supported by Prince Sultan University}
\thanks{$^{1}$Prince Sultan University, Saudi Arabia 
        {\tt\small bbenjdira@psu.edu.sa}}%
\thanks{$^{2}$Prince Sultan University, Saudi Arabia 
        {\tt\small 215110375@psu.edu.sa}}%
\thanks{$^{3}$Prince Sultan University, Saudi Arabia/Gaitech Robotics, China/CISTER, INESC-TEC, ISEP, Polytechnic Institute of Porto, Portugal
        {\tt\small akoubaa@psu.edu.sa}}%
\thanks{$^{4}$Al-Imam Mohamed bin Saud University, Saudi Arabia
        {\tt\small adel.ammar@ccis.imamu.edu.sa}}%
\thanks{$^{5}$Research Laboratory Smart Electricity \& ICT, SEICT, LR18ES44. National Engineering School of Carthage, University of Carthage, Tunisia
        {\tt\small }}%
}

\begin{document}

\maketitle
\thispagestyle{empty}
\pagestyle{empty}

\begin{abstract}

Unmanned Aerial Vehicles are increasingly being used in surveillance and traffic monitoring thanks to their high mobility and ability to cover areas at different altitudes and locations. One of the major challenges is to use aerial images to accurately detect cars and count-them in real-time for traffic monitoring purposes. Several deep learning techniques were recently proposed based on convolution neural network (CNN) for real-time classification and recognition in computer vision. However, their performance depends on the scenarios where they are used. In this paper, we investigate the performance of two state-of-the art CNN algorithms, namely Faster R-CNN and YOLOv3, in the context of car detection from aerial images. We trained and tested these two models on a large car dataset taken from UAVs. We demonstrated in this paper that YOLOv3 outperforms Faster R-CNN in sensitivity and processing time, although they are comparable in the precision metric. 


\end{abstract}

\section{INTRODUCTION}

Unmanned aerial vehicles (UAVs) are being more and more adopted in surveillance and monitoring tasks due to their flexibility and  great mobility. UAVs produce high resolution images for wide fields of view in real time. The adoption of UAVs have been inhibited first by the low accuracy of object detection algorithms based on the traditional approaches of machine learning.  However, since the emergence of deep learning algorithms and especially the convolution neural networks, object detection and recognition have shown a notable increase of accuracy. This paves the way towards a widespread adoption of UAVs for data acquisition and analysis in many engineering fields. This is why it is estimated that UAV global sales are expected to surpass \$12 billion by 2021 \cite{businessinsider}. 

UAVs have enabled a large variety of applications, such as tracking \cite{dronetrack, Khan2017}, surveillance \cite{Ding2018} and in particular mapping and land surveying \cite{Tariq2016}. They are also used in surveillance applications given their ability to cover open areas at different altitudes and provide high-resolution videos and images. In this paper, we consider the scenario of vehicle surveillance and traffic monitoring, where a drone is used to detect and count vehicles from aerial video streams.  

With advances of deep learning, and in particular convolution neural network (CNN), in computer vision applications, the accuracy of classification and object recognition has reached an impressive improvement. The evolution of Graphic Processing Units (GPUs) also significantly contributed to the adoption of CNN in computer vision overcoming the problems of real-time processing of computation intensive tasks through parallelization. In addition, latest trends in cloud robotics \cite{dp2019, survey2016,ICARSC2017} have also enabled offloading heavy computations, such as video stream analysis, to the cloud. This allows to process video streams in real-time using advanced deep learning algorithms in the context of surveillance applications. 

Since 2012, several CNN algorithms and architectures were proposed such as YOLO  and its variants \cite{YOLO2016, YOLOv2,YOLOv3}, R-CNN and its variants \cite{R-CNN, Fast_R-CNN, Faster_R-CNN_conf, Faster_R-CNN_journal}. R-CNN is a region-based CNN, proposed by Girshick et al.\cite{R-CNN},  which combines region-proposals algorithm with CNN. The idea is to extract 2000 regions through a selective search, then instead of working on the whole image, the classification will occur on the selected regions. The same authors improved their algorithm by overcoming the limitation of R-CNN that consists in generating a convolutional feature map where is the input is the image instead of the regions. Region of proposals are then identified from the convolutional feature map. Then, Shaoqing Ren et al. \cite{Faster_R-CNN_conf}  proposed Faster R-CNN by replacing the selective search of region, which is slow, by an object detection algorithm. 

On the other hand, in 2016, YOLO was proposed by Joseph Redmon using a different approach named: You Only Look Once\cite{YOLO2016}. Unlike region-based approaches, YOLO passes the n by n image only once  in a fully convolutional neural network (FCNN), which makes it quite fast and real-time. It splits the image into grids of dimension m by m, and generates bounding boxes and their class probabilities. YOLOv2\cite{YOLOv2} overcomes the relatively high localization error and low recall (measure of how good is the localization of all objects), as compared to region-based techniques,  by making batch-normalization and higher resolution classifier. Recently, in 2018, YOLOv3\cite{YOLOv3} is released and is characterized by a higher accuracy and replaces softmax function with logistic regression and threshold. 

In this paper, we consider the performance evaluation of these two categories of CNN architectures in the context of car detection from aerial images, in terms of accuracy and processing time. We consider the latest approaches of the two categories, namely Faster R-CNN for region-based algorithms, and YOLOv3.  

The rest of the paper is organized as follows: Section 2 discusses related works about car detection from UAV imagery. Section 3 provides an overview of Faster R-CNN model and the YOLOv3 model, and discusses a theoretical comparison between them. Section 4 presents the performance evaluation of the algorithms for car detection from aerial images. Section 5 concludes the paper and discusses the main results. 
\section{RELATED WORKS}

In this section, we present an overview of the main works related to car detection problem using Convolutional Neural Networks. 

Chen et al. \cite{Chen2014} proposed a new model named hybrid deep neural network (HDNN). This model is based on sliding windows and deep CNN. The key idea of the model was to replicate the convolutional layers at different scales to make the model able to recognize cars at different scales. They used a modified sliding-window search that is able to center sliding-windows around cars.  Although the originality of the idea and the improved car detection rate compared to other solutions at the time, their approach is highly time-consuming as it needs about 7 seconds to process one image even using GPU acceleration. 

Ammour et al. \cite{Ammour2017} used two-stage method for the car detection problem. The first phase is the candidate region extraction stage and uses mean-shift algorithm  to segment the image. The second phase is the car detection stage that uses the VGG16 [15] model to extract region feature, followed by  a Support Vector Machine (SVM) classifier that uses this feature to classify it if it is car or non-car. Although the contribution surpasses competitors in terms of accuracy but it is still time-consuming and could not be used for real time applications. Indeed, the algorithm takes around 12 minutes to process 3456*5184 image. This is due to the different stages that the model uses (mean-shift \cite{meanshift} segmentation, VGG16 \cite{vgg} feature extraction, SVM classification). The main computation load is resulted from the mean-shift segmentation, which is their core contribution for object localization. It is comparable to the R-CNN approach where the algorithm suggested for the object localization is the region proposal algorithm and both suffer from the computation load due to the object localization problem. We will show in the next section how this problem is solved by Faster R-CNN\cite{Faster_R-CNN_conf,Faster_R-CNN_journal} and YOLOv3\cite{YOLOv3}.

In this paper, we consider Faster R-CNN and YOLOv3, which are the state of the art algorithms of CNN for object detection. We selected them due to their excellent performance and our objective is to compare between them in the context of the car detection problem. In this next section, we will present a theoretical overview of the two approaches. 

\section{ Theoretical Overview Faster R-CNN and YOLOv3}
Faster R-CNN and YOLOv3 are the state of the art algorithms used for generic object detection and were successfully adapted to many recognition problems. This paper aims to make a deeper look at the differences between these two algorithms and precisely the use of these algorithms for the car detection problem.
\subsection{Faster R-CNN}
The Faster R-CNN model is divided into two modules: the region proposal network (RPN) and a Fast R-CNN detector. RPN is a fully convolutional network used the generate region proposals with multiple scales and aspect ratios serving as an input for the second module. Region proposals are the bounding boxes inside the input image which possibly contain the candidate objects. The RPN and the Fast R-CNN detector share the same convolutional layers. Faster R-CNN, by consequence, could be considered as a single and a unified network for object detection. To generate high quality object proposal, we can use a highly descriptive feature extractor (VGG16 \cite{vgg} for example) in the convolutional layers. 
\begin{figure}  
\begin{center}  
\includegraphics[width=8cm]{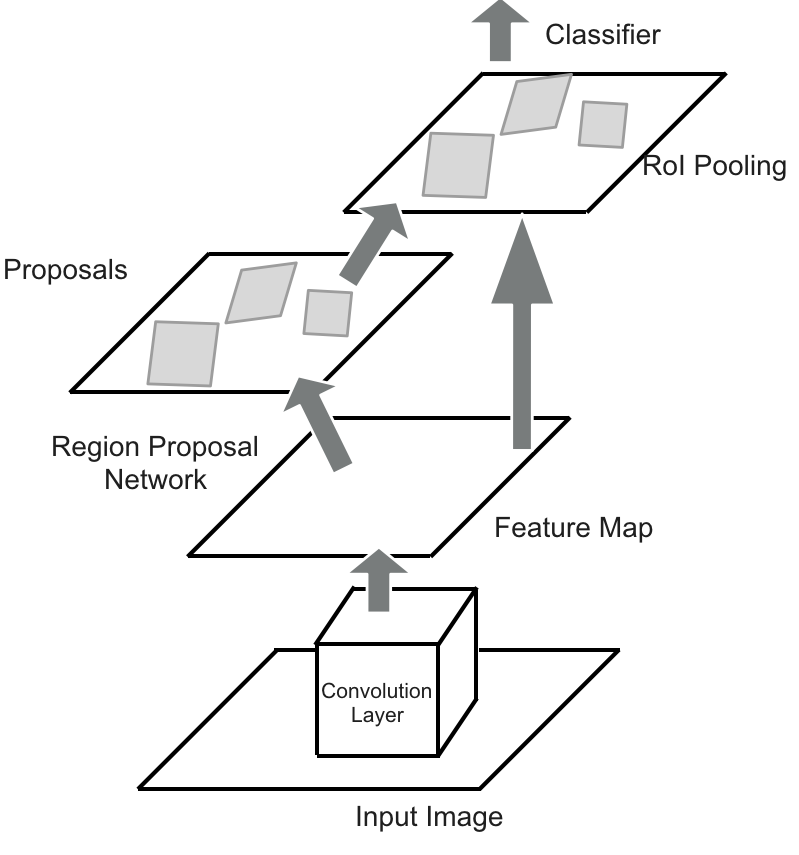}
\caption{\small \sl Faster R-CNN architecture\label{fig:Stupendous}}  
\end{center}  
\end{figure} 
The Fast R-CNN detector uses as input many regions of interest (ROIs). Then, the ROI pooling layer extracts for each ROI a feature vector. This feature vector will constitute the input for a classifier formed by a series of fully connected (FC) layers. Finally, we get two outputs. The first output is a sequence of probabilities estimated over the different objects considered. In our case we will have the probabilities of the classes “car” and “background”. The second output is the coordinates of the bounding-box (bbox) values. 
Concerning the RPN, it generates from the UAV image a list of bounding boxes. Each one is associated with an objectness score. The objectness measures membership of the selected portion of the image to a set of object classes versus background\cite{Faster_R-CNN_journal}. 
In this paper, the Inception ResNet v2 \cite{ResNetv2} model is used as the shared convolutional network in Faster R-CNN.  

\subsection{Architecture of YOLOv3}
 YOLOv3 \cite{YOLOv3} is an improvement made over its predecessors: YOLO v1 \cite{YOLO2016} and YOLO v2 \cite{YOLOv2} (named also YOLO9000).
 \subsubsection{Description of YOLO v1 algorithm}
 YOLO contains 24 convolutional layers followed by 2 fully connected layers. Some convolutional layers use convolutions of size  1×1  to reduce depth dimension of the feature maps. A faster version of YOLO, named Fast YOLO, uses only 9 convolutional layers but this impacts the accuracy. The general architecture is displayed on Fig 2.\par
 YOLO divides the input image into an \(S\times S\) grid. A grid cell can only be associated to one object and it can only predict a fixed number B of boundary boxes, each box is associated with a one box confidence score. So, the information to be predicted for each bounding box contains 5 values\( (x,y,w,h,box\_confidence\_score)\). Concerning the detected object, the grid cell will be associated to a sequence of class probabilities to estimate the classification of the object over the C classes of the model. The central concept of YOLO v1 was to build a single CNN Network to predict a tensor of dimensions:

\begin{center}
    \( S\times S\times (B*5+C) \) \par
\end{center} \par
\( S\times S\): is the number of the grid cells of the system \par
\( B\): is the number of the bounding boxes per grid cell. \par
\(C\): is the number of the classes we train our network on.     
For evaluating the YOLO on PASCAL VOC \cite{PascalVOC}, they used for values \cite{YOLO2016}: S=7 ,B=2, C=20 (as PASCAL VOC \cite{PascalVOC} has 20 classes of objects). We get finally a (7,7,30) tensor. In the final prediction, we keep only high box confidence scores and the object class with highest probability. 
Indeed, this the major contribution made by YOLO over the existing CNN architectures: to design a CNN network to predict \(S \times S \times (B*5+C) \)  tensor. 
\begin{figure*}  
\begin{center}  
\includegraphics[width=13cm]{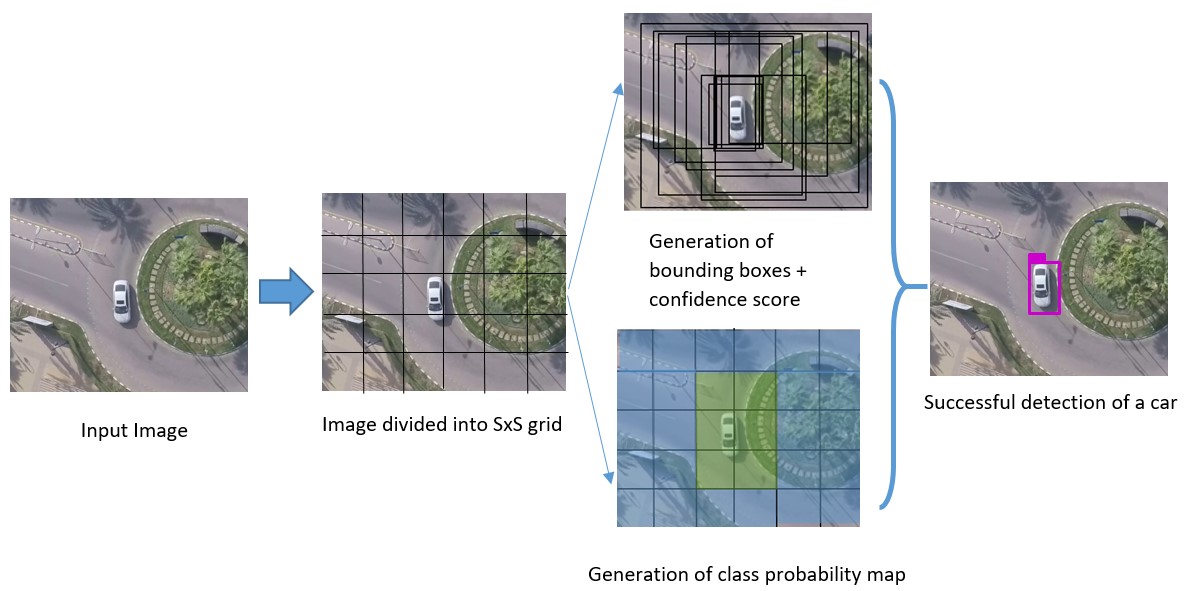}
\caption{\small \sl YOLO v3 Architecture }\label{fig:Stupendous}  
\end{center}  
\end{figure*} \par
To choose the right bounding box for the grid cell, we select the one with the highest IoU (intersection over union) with the ground truth. To calculate loss, YOLO uses sum-squared error between the predictions and the ground truth. The total loss function is composed of three loss functions: the confidence loss (the objectness of the box), the localization loss (the sum-squared error between the predictions and the ground truth) and the classification loss (the squared error of the class conditional probabilities for each class). \par
To remove duplicate detections for the same object, YOLO uses non-maximal suppression. If we have \(IoU \geq 0.5 \) between any of the predictions in the image, non-maximal suppression delete the prediction with the lowest confidence score. \par
When introduced, YOLO outperforms other CNN architectures in term of speed, keeping or outperforming the state of the art mAP (mean Average Precision). Although it makes more localization errors but it is less likely to predict false positives. It outperforms state of the art methods in generalization. \par
The training of YOLO is composed of 2 phases. First, we train a  classifier network like VGG16 \cite{vgg} . Like all of the state of the art methods, the classifier network is pre-trained on ImageNet using image input at \(224 \times 224 \). Secondly, we replace the fully connected layers with a convolution layer and make a complete training from end to end for the object detection. 
\subsubsection{Improvements made in YOLO v2}
A new competitor for YOLO is appeared, the SSD \cite{SSD} (Single Shot MultiBox Detector). This algorithm outperforms YOLO in accuracy for real-time object detection. Thus, YOLO v2 is introduced applying many improvements to increase accuracy and processing time. \par
First improvement made is the Batch Normalization  (BN) \cite{BN} technique, introduced in 2015. It is used to normalize the input layers by adjusting and scaling the activations. By adding batch normalization on all convolutional layers in YOLO, mAP is improved by 2\%. Also, using BN, Dropout \cite{dropout} technique can be removed from the model without having overfitting. \par
Second improvement made is the use of High Resolution Classifier. The  size \(224 \times 224 \)of the input image in the first phase of YOLO training is replaced by the size \(448 \times 448\). This makes the detector working better on higher resolutions and increasing the mAP by 4\%. \par
Third improvement made is the use of convolutional with anchors boxes. The fully connected layers responsible for predicting the boundary box is removed and we move the class prediction form the grid cell level to the boundary box level. The adoption of the anchor boxes makes a slight decrease in the mAP by 0.3 \% but improves the recall from 81\% to 88\% increasing the chance to detect all the ground truth objects. \par
The fourth improvement made is the use of dimension clusters. We use the K-means clustering on the training set bounding boxes to automatically find the best anchors boxes. Instead of the Euclidean distance, the IoU scores are used for the clustering. \par
The fifth improvement made is the direct location prediction. The predictions are made on the offsets to the anchors. We predict five parameters \((t\textsubscript{x},t\textsubscript{y},t\textsubscript{w},t\textsubscript{h},t\textsubscript{o})\) and then apply a function σ  to predict the bounding box. This makes the network more stable and easier to learn. The fourth and the fifth improvements increase the mAP by 5\%. \par
The sixth improvement made is the fine-grained features. To improve the capability of detecting small objects, YOLO adopts an approach named pass-through layer. This concatenates the high resolution features with the low resolution features, similar to the identity mapping in ResNet \cite{ResNet}. This improve the mAP with 1\%. \par
The seventh improvement made is the Multi-scale training. The YOLO v2 uses an input resolution of \(448 \times 448\). After adding of anchor boxes, resolution is changed to \(416 \times 416\). Instead of fixing the input image size, every 10 batches the network randomly chooses a new image dimension size. This helps to predict well across a variety of input dimensions. \par
\subsubsection{Improvements made in YOLO v3}
The first improvement made with YOLOv3 is the use of the multi-label classification, which is different from the mutual exclusive labeling used in the previous versions. It uses a logistic classifier to calculate the likeliness of the object being of a specific label. Previous versions use the softmax function to generate the probabilities form the scores. For the classification loss, it uses the binary cross-entropy loss for each label, instead of the general mean square error used in the previous versions. \par
The second improvement made is the use of different bounding box prediction. It associates the objectness score 1 to the bounding box anchor which overlaps a ground truth object more than others. It ignores others anchors that overlaps the ground truth object by more than a chosen threshold (0.7 is used in the implementation). Therefore, YOLOv3 assigns one bounding box anchor for each ground truth object. \par
The third improvement made is the use of prediction across scales using the concept of feature pyramid networks. YOLOv3 predicts boxes at 3 different scales and then extracts features from those scales. The prediction result of the network is a 3-d tensor that encodes bounding box, objectness score and prediction over classes. This is why the tensor dimensions at the end are changed from previous versions to:
\begin{center}
\(N \times N \times (3*(4+1+C))\) \par
\end{center} \par
\(N \times N\): is the number of the grid cells of the system \par
\(3\): to decode the features extracted from each of the 3 scales \par 
\(4+1\): to decode the bounding boxes offsets + objectness score \par
\(C\): is the number of the classes we train our network on. \par
This allows to get better semantic information from the up-sampled features and finer-grained information from the earlier feature map. \par  
The fifth improvement made is the new CNN feature extractor named Darknet-53. It is a 53 layered CNN that uses skip connections network inspired from ResNet \cite{ResNet}. It uses also \(3 \times 3\) and \(1 \times 1\) convolutional layers. It has shown the state of the art accuracy but with fewer floating point operations and better speed. For example, it has less floating point operations than ResNet-152 but the same performance at a double speed. 
 
\section{ Experimental comparison between Faster R-CNN and YOLOv3}
In this section we will describe the dataset used (training set and test set). We will specify the used hardware and software in our experiments. The evaluation of the algorithms is based on five metrics described below. The video demonstration of a real time car detection from UAV is available at\cite{demo-video-link}. A screenchot of the application of Faster R-CNN and Yolo v3 on UAV images are shown on Fig 3 and Fig 4 respectively. 
\subsection{Description of the Dataset}
To perform the experimental part of our study, we built a UAV imagery dataset divided into a training set and a test set. The training set contains 218 images and 3,365 instances of labeled cars. The test set contains 52 images and 737 instances of cars. This dataset was collected from images taken by an UAV flown above Prince Sultan University campus and from an open source dataset available in Github \cite{aerial-car-dataset}. We tried to collect cars from different environments and scales to assure the validity of our experiment  and to test the genericity of the algorithms. For example, some images are taken from an altitude of 55m and others are taken from above 80m. 
\begin{figure}  
\begin{center}  
\includegraphics[width=8cm]{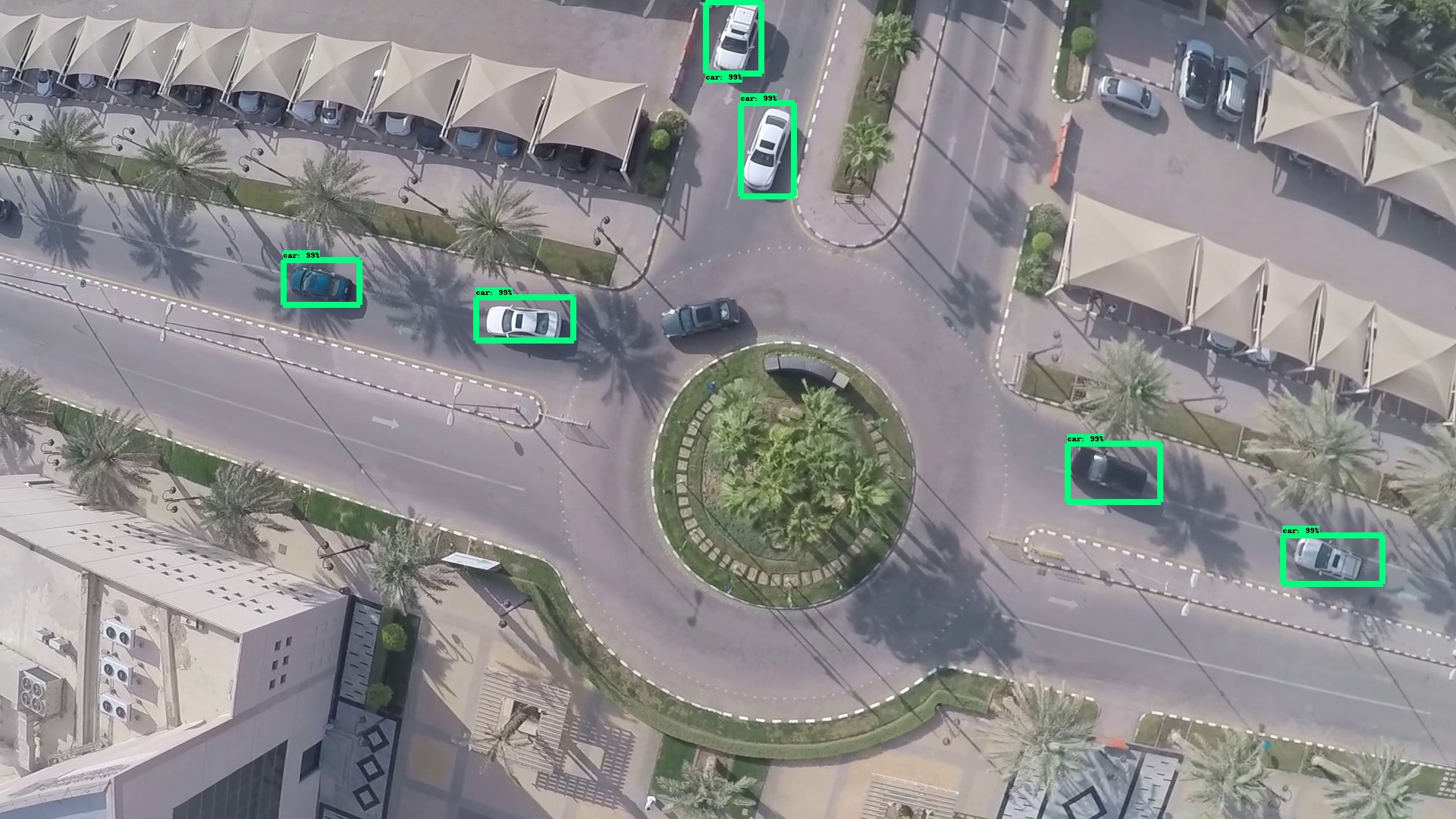}
\caption{\small \sl Car detection using Faster R-CNN\label{fig:Stupendous}}  
\end{center}  
\end{figure} 
\begin{figure}  
\begin{center}  
\includegraphics[width=8cm]{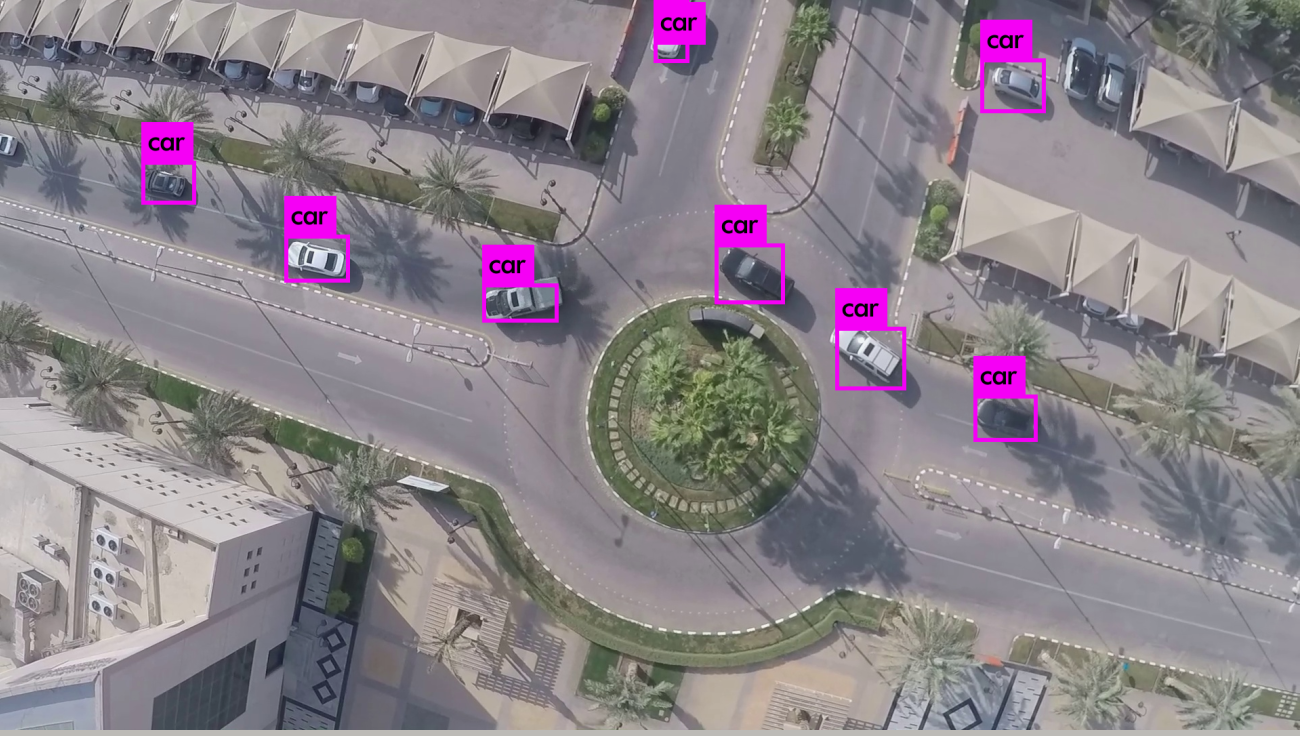}
\caption{\small \sl Car detection using Yolo v3\label{fig:Stupendous}}  
\end{center}  
\end{figure} 
\subsection{Description of the Hardware and Software tools}
Concerning the training of Faster R-CNN \cite{Faster_R-CNN_conf, Faster_R-CNN_journal}, we  used as software the Tensorflow Object Detection API \cite{tensorflow-oda}. We chose the provided model of Faster R-CNN with convolutional backend the Inception ResNet v2 CNN network. We optimized the training of Faster R-CNN  using stochastic gradient descent with momentum set to 0.89. The learning rate is set to 0.00019. We trained for 200K steps. In the preprossessing phase, we set for image resizing operation, the minimal dimension to 600 and the maximal dimension to 1024. For data augmentation, we used only a random horizontal flip operation among the training set. We set the batch size to one. For the specific parameters belonging to Faster R-CNN itself, we set the value 300 as maximal object proposal in total and 100 as maximal object proposal per class. The algorithm is trained to recognize only one class which is the class "car". \par
Concerning the training of YOLO v3, we used the YOLO v3 provided code \cite{YOLOv3}. Concerning the parameters, we used the yolov3 default configuration. We optimized the training using stochastic gradient descent with momentum set to 0.9. The learning rate is set to 0.001 and the weight decay is set to 0.005. The value for both height and width is set to 608. The batch size is set to 64. Concerning the YOLO v3 parameters, the input image is subdivided to 16*16 grids. Anchors that overlaps the ground truth object by less than a threshold value (0.7) are ignored. \par
Concerning the configurations of the computer used in this research, they are: \par
\begin{itemize}
\item CPU: Intel Core i9-8950HK (six cores, Coffee Lake architecture)
\item Graphic card: Nvidia GTX 1080, 8GB GDDR5
\item RAM: 32 GB RAM
\item Operating system: Linux (Ubuntu 16.04)
\end{itemize}
\subsection{Performance evaluation and metrics}
To compare the performance between the two algorithms, we have used five parameters (Precision, Recall, F1 Score, Quality and processing speed). The four first parameters are defined below: \par
\begin{itemize}
\item \(Precision = \frac{TP}{TP+FP} \)
\item \(Recall=Sensitivity= \frac{TP}{TP+FN} \)
\item \(F1 Score= 2*Pecision*Recall (Precision + Recall)\)
\item \(Quality=\frac{TP}{TP+FP+FN}\)
\end{itemize}
Where TP (True Positives) indicates the number of cars successfully detected by the algorithm. FP (False Positives) indicates the number of non-car objects that are falsely detected as cars. FN (False Negative) indicates the number of cars that the algorithm did not recognize them as cars. 
\subsection{Comparison between Faster R-CNN and YOLO v3}
Here we will try to evaluate both of the algorithms based on the five metrics we identified. Table \ref{table1} contains all the values measured for each algorithm. 

\begin{table}[]
\begin{tabular}{lll}
\hline
\textbf{Measure}                         & \textbf{\begin{tabular}[c]{@{}l@{}}Faster R-CNN\\   (test dataset)\end{tabular}} & \textbf{\begin{tabular}[c]{@{}l@{}}YOLOv3 \\   (test dataset)\end{tabular}} \\ \hline
\textbf{TP (True positives)}             & 578                                                                              & 751                                                                         \\ \hline
\textbf{FP (False positives)}            & 2                                                                                & 2                                                                           \\ \hline
\textbf{FN (False negatives)}            & 150                                                                              & 7                                                                           \\ \hline
\textbf{Precision (TPR)}                 & 99.66\%                                                                          & 99.73\%                                                                     \\ \hline
\textbf{Sensitivity (recall)}            & 79.40\%                                                                          & 99.07\%                                                                     \\ \hline
\textbf{F1 Score}                        & 88.38\%                                                                          & 99.94\%                                                                     \\ \hline
\textbf{Quality}                         & 79.17\%                                                                          & 98.81\%                                                                     \\ \hline
\textbf{Processing time (Av. in ms)} & 1.39 s                                                                            & 0.057 ms                                                                      \\ \hline
\end{tabular}
\caption{Evaluation metrics of Faster R-CNN and YOLOv3 }
\label{table1}
\end{table}
The evaluation metrics show that both of the algorithms has high precision rate (99.66\% for Faster R-CNN vs 99.73\% for YOLOv3). This high value indicates that when they classify an object as car, it is very highly probable that this object is a car. So, the ability of the algorithms to detect true cars is very high. Percentage that the algorithms classify non-car objects as car is very rare (0.34\% for Faster R-CNN versus 0.27\% for YOLOv3). But when comparing the recall, we note that YOLOv3 outperforms clearly Faster R-CNN (79.40\% for Faster R-CNN versus 99.07\% YOLOv3). The recall measures the ability of the algorithm to detect all the instances of cars in the image. YOLOv3 is more capable to extract all the instances of cars in one image, Faster R-CNN misses some instances more than YOLOv3. This is due to the high number of FN (False Negatives) which is 150 for Faster R-CNN versus 2 for YOLOv3. Considering F1, which is a harmonic average of the precision and recall that gives a global idea about robustness of the algorithm (his ability to extract all the instances of cars and to not falsely extract non-car objects). We note here that YOLOv3 outperforms Faster R-CNN as it has higher recall. The quality measure is a similar metric to measure the robustness of the algorithm. This measures indicates similarly the robustness of YOLOv3 versus Faster R-CNN. \par
Concerning the processing time, we measured the processing time for one-time detection (detection of number of cars in one image) for 15 sizes of input image (form the size of 100px*100px to 1500px*1500px, increasing by 100 pixels at each size). For YOLOv3, the processing time for each image ranges between 0.056 ms and 0.060 ms with an average of 0.057 ms. For Faster R-CNN, the processing time for each image ranges between 1.18 s and 2.90 s with an average of 1.39 s. This shows the great performance gap between the two algorithms in processing one image per time. We noted also that the processing time does not depend on the size of the image. The average is the same independently from the size.
\section{Conclusion}
In this paper, we have made a comparison between the state of the art algorithms for object detection (Faster R-CNN and YOLOv3). We made an experimental comparison between them in car detection task from UAV images. We began by a theoretical description of both algorithms, citing the architectural design and the improvements that precedes the current design of them. Then, we made an experimental comparison using a labeled car dataset divided into a train dataset and a test dataset. This dataset is used to train both models and test their performance. The performance evaluation is performance metrics is made using five metrics: precision, recall, F1 score, quality and processing time. Based on the results obtained, we first found that both algorithms are comparable in precision, which mean that both of them have high capability to correctly classify car object in the image. But we found also that YOLO v3 outperforms Faster R-CNN in sensitivity which mean that YOLO V3 is more capable to extract all the cars in the image with 99.07\% accuracy. Concerning the processing time for one image detection, we found also that YOLOv3 outperforms Faster R-CNN. This study serves as a guidance for traffic monitors that plan to use UAVs for traffic monitoring and demonstrates that YOLOv3 can be used for traffic monitoring in UAV imagery. Also, it serves for researchers that need to choose the best algorithm for object detection according to their needs. \par
However, our study be extended for general vehicle detection (bicycle, motorcycle, bus, truck…). Besides, the dataset can be extended to add different lighting conditions (day, night, morning, evening) and different environmental factors (urban, rural, crowded traffic, winter, summer…). 

\section*{Acknowledgments}
This work is supported by the Robotics and Internet of Things Lab of Prince Sultan University.

\bibliographystyle{ieeetr}
\bibliography{biblio}

\begin{thebibliography}{10}

\bibitem{businessinsider}
``{Business Insider. Available online: goo.gl/uUJRWD (accessed on 8 April
  18).}.''

\bibitem{dronetrack}
A.~Koubaa and B.~Qureshi, ``Dronetrack: Cloud-based real-time object tracking
  using unmanned aerial vehicles over the internet,'' {\em {IEEE} Access},
  vol.~6, pp.~13810--13824, 2018.

\bibitem{Khan2017}
M.~Khan, K.~Heurtefeux, A.~Mohamed, K.~A. Harras, and M.~M. Hassan, ``Mobile
  target coverage and tracking on drone-be-gone uav cyber-physical testbed,''
  {\em IEEE Systems Journal}, vol.~12, pp.~3485--3496, Dec 2018.

\bibitem{Ding2018}
G.~Ding, Q.~Wu, L.~Zhang, Y.~Lin, T.~A. Tsiftsis, and Y.~Yao, ``An amateur
  drone surveillance system based on the cognitive internet of things,'' {\em
  IEEE Communications Magazine}, vol.~56, pp.~29--35, Jan 2018.

\bibitem{Tariq2016}
A.~Tariq, S.~M. Osama, and A.~Gillani, ``Development of a low cost and light
  weight uav for photogrammetry and precision land mapping using aerial
  imagery,'' in {\em 2016 International Conference on Frontiers of Information
  Technology (FIT)}, pp.~360--364, Dec 2016.

\bibitem{dp2019}
A.~Koubâa, B.~Qureshi, M.-F. Sriti, A.~Allouch, Y.~Javed, M.~Alajlan,
  O.~Cheikhrouhou, M.~Khalgui, and E.~Tovar, ``Dronemap planner: A
  service-oriented cloud-based management system for the internet-of-drones,''
  {\em Ad Hoc Networks}, vol.~86, pp.~46 -- 62, 2019.

\bibitem{survey2016}
R.~Chaari, F.~Ellouze, A.~Koubaa, B.~Qureshi, N.~Pereira, H.~Youssef, and
  E.~Tovar, ``Cyber-physical systems clouds: {A} survey,'' {\em Computer
  Networks}, vol.~108, pp.~260--278, 2016.

\bibitem{ICARSC2017}
A.~Koubaa, B.~Qureshi, M.~Sriti, Y.~Javed, and E.~Tovar, ``A service-oriented
  cloud-based management system for the internet-of-drones,'' in {\em 2017
  {IEEE} International Conference on Autonomous Robot Systems and Competitions,
  {ICARSC} 2017, Coimbra, Portugal, April 26-28, 2017}, pp.~329--335, 2017.

\bibitem{YOLO2016}
J.~Redmon, S.~K. Divvala, R.~B. Girshick, and A.~Farhadi, ``You only look once:
  Unified, real-time object detection,'' in {\em 2016 {IEEE} Conference on
  Computer Vision and Pattern Recognition, {CVPR} 2016, Las Vegas, NV, USA,
  June 27-30, 2016}, pp.~779--788, 2016.

\bibitem{YOLOv2}
J.~Redmon and A.~Farhadi, ``{YOLO9000:} better, faster, stronger,'' in {\em
  2017 {IEEE} Conference on Computer Vision and Pattern Recognition, {CVPR}
  2017, Honolulu, HI, USA, July 21-26, 2017}, pp.~6517--6525, 2017.

\bibitem{YOLOv3}
J.~Redmon and A.~Farhadi, ``Yolov3: An incremental improvement,'' {\em CoRR},
  vol.~abs/1804.02767, 2018.

\bibitem{R-CNN}
R.~Girshick, J.~Donahue, T.~Darrell, and J.~Malik, ``{Rich feature hierarchies
  for accurate object detection and semantic segmentation},'' in {\em
  Proceedings of the IEEE Computer Society Conference on Computer Vision and
  Pattern Recognition}, pp.~580--587, 2014.

\bibitem{Fast_R-CNN}
R.~Girshick, ``{Fast R-CNN},'' in {\em Proceedings of the IEEE International
  Conference on Computer Vision}, 2015.

\bibitem{Faster_R-CNN_conf}
R.~Girshick, ``{Fast R-CNN},'' in {\em Proceedings of the IEEE International
  Conference on Computer Vision}, 2015.

\bibitem{Faster_R-CNN_journal}
S.~Ren, K.~He, R.~Girshick, and J.~Sun, ``{Faster R-CNN: Towards Real-Time
  Object Detection with},'' {\em IEEE TRANSACTIONS ON PATTERN ANALYSIS AND
  MACHINE INTELLIGENCE}, 2017.

\bibitem{Chen2014}
X.~Y. Chen, S.~M. Xiang, C.~L. Liu, and C.~H. Pan, ``{Vehicle Detection in
  Satellite Images by Hybrid Deep Convolutional Neural Networks},'' {\em Ieee
  Geoscience and Remote Sensing Letters}, 2014.

\bibitem{Ammour2017}
N.~Ammour, H.~Alhichri, Y.~Bazi, B.~Benjdira, N.~Alajlan, and M.~Zuair, ``{Deep
  Learning Approach for Car Detection in UAV Imagery},'' {\em Remote Sensing},
  vol.~9, p.~312, mar 2017.

\bibitem{meanshift}
D.~Comaniciu and P.~Meer, ``{Mean shift: A robust approach toward feature space
  analysis},'' {\em IEEE Transactions on Pattern Analysis and Machine
  Intelligence}, 2002.

\bibitem{vgg}
K.~Simonyan and A.~Zisserman, ``{Very Deep Convolutional Networks for
  Large-Scale Image Recognition},'' {\em International Conference on Learning
  Representations (ICRL)}, 2015.

\bibitem{ResNetv2}
C.~Szegedy, S.~Ioffe, V.~Vanhoucke, and A.~Alemi, ``{Inception-v4,
  Inception-ResNet and the Impact of Residual Connections on Learning},'' feb
  2016.

\bibitem{PascalVOC}
M.~Everingham, S.~M. Eslami, L.~{Van Gool}, C.~K. Williams, J.~Winn, and
  A.~Zisserman, ``{The Pascal Visual Object Classes Challenge: A
  Retrospective},'' {\em International Journal of Computer Vision}, 2014.

\bibitem{SSD}
W.~Liu, D.~Anguelov, D.~Erhan, C.~Szegedy, S.~Reed, C.~Y. Fu, and A.~C. Berg,
  ``{SSD: Single shot multibox detector},'' in {\em Lecture Notes in Computer
  Science (including subseries Lecture Notes in Artificial Intelligence and
  Lecture Notes in Bioinformatics)}, 2016.

\bibitem{BN}
S.~Ioffe and C.~Szegedy, ``{Batch Normalization: Accelerating Deep Network
  Training by Reducing Internal Covariate Shift},'' feb 2015.

\bibitem{dropout}
N.~Srivastava, G.~Hinton, A.~Krizhevsky, and R.~Salakhutdinov, ``{Dropout: A
  Simple Way to Prevent Neural Networks from Overfitting},'' tech. rep., 2014.

\bibitem{ResNet}
K.~He, X.~Zhang, S.~Ren, and J.~Sun, ``{Deep Residual Learning for Image
  Recognition},'' {\em Arxiv.Org}, 2015.

\bibitem{demo-video-link}
``Video demonstration of real time car detection from uav images, [available
  online at: https://www.youtube.com/watch?v=rlpuhjmkcv4], [accessed on
  18-12-2018].''

\bibitem{aerial-car-dataset}
``{Aerial-car-dataset, available online on:
  https://github.com/jekhor/aerial-cars-dataset, accessed on (16-10-2018)}.''

\bibitem{tensorflow-oda}
``{Tensorflow-object-detection-api, available online on:
  https://github.com/tensorflow/models/tree/master/research/object\_detection,
  accessed on (16-10-2018)}.''

\end{thebibliography}

\end{document}